\documentclass[letterpaper, 10 pt, conference]{ieeeconf}  

\IEEEoverridecommandlockouts                              
\usepackage[T1]{fontenc} 
\usepackage{booktabs} 
\usepackage{hyperref}
\usepackage{graphicx} 
\usepackage{subfigure}
\usepackage{multirow} 
\usepackage{makecell} 
\usepackage{amsfonts} 
\usepackage[sorting=none, style=ieee]{biblatex}
\usepackage[linesnumbered,ruled,commentsnumbered]{algorithm2e}
\usepackage{amsmath} 
\title{\LARGE \bf
GDTS: Goal-Guided Diffusion Model with Tree Sampling for Multi-Modal Pedestrian Trajectory Prediction
}

\author{Ge Sun, Sheng Wang, Lei Zhu, Ming Liu, and Jun Ma, \textit{Senior Member, IEEE}
\thanks{Ge Sun and Sheng Wang are with the Division of Emerging Interdisciplinary Areas, The Hong Kong University of Science and Technology, Hong Kong SAR, China. {\tt\small gsunah@connect.ust.hk, swangei@connect.ust.hk}}%
\thanks{Ming Liu is with Robotics and Autonomous Systems Thrust, The Hong Kong University of Science and Technology (Guangzhou), Guangzhou, China. {\tt\small eelium@hkust-gz.edu.cn}}%
\thanks{Lei Zhu and Jun Ma are with the Robotics and Autonomous Systems Thrust, The Hong Kong University of Science and Technology (Guangzhou), Guangzhou, China, and also with the Department of Electronic and Computer Engineering, The Hong Kong University of Science and Technology, Hong Kong SAR, China. {\tt\small leizhu@ust.hk, jun.ma@ust.hk}}%
}

\begin{document}

\newcommand{\etal}{\textit{et al.}} 

\maketitle
\thispagestyle{empty}
\pagestyle{empty}

\begin{abstract}

Accurate prediction of pedestrian trajectories is crucial for improving the safety of autonomous driving. However, this task is generally nontrivial due to the inherent stochasticity of human motion, which naturally requires the predictor to generate multi-modal prediction. Previous works leverage various generative methods, such as GAN and VAE, for pedestrian trajectory prediction. Nevertheless, these methods may suffer from mode collapse and relatively low-quality results. The denoising diffusion probabilistic model (DDPM) has recently been applied to trajectory prediction due to its simple training process and powerful reconstruction ability. However, current diffusion-based methods do not fully utilize input information and usually require many denoising iterations that lead to a long inference time or an additional network for initialization.
To address these challenges and facilitate the use of diffusion models in multi-modal trajectory prediction, we propose GDTS, a novel Goal-Guided Diffusion Model with Tree Sampling for multi-modal trajectory prediction. Considering the "goal-driven" characteristics of human motion, GDTS leverages goal estimation to guide the generation of the diffusion network. A two-stage tree sampling algorithm is presented, which leverages common features to reduce the inference time and improve accuracy for multi-modal prediction. Experimental results demonstrate that our proposed framework achieves comparable state-of-the-art performance with real-time inference speed in public datasets.

\end{abstract}

\section{INTRODUCTION}

Time-series forecasting has been widely applied in various fields, including finance~\cite{financial_2020}, climate~\cite{climate_2020}, and healthcare~\cite{healthcare_2021}. With the rapid development of autonomous driving techniques, trajectory prediction, which could be formulated as a time-series forecasting problem, has also gained great attention~\cite{SocialGAN_2018_CVPR, Chen_2022_IROS, Sheng_2024_ICRA}. Accurate trajectory prediction of surrounding agents can enhance the ability of autonomous systems to handle interactions, thereby improving the safety and effectiveness of the entire system.
Many trajectory prediction datasets categorize traffic agents into multiple classes based on their properties~\cite{Nuscenes_2020_CVPR, WOMD_2021_ICCV}, with pedestrians also being separated as a distinct class. Since pedestrians are relatively more vulnerable when interacting with other classes of traffic agents, special attention is required. Thus, in this paper, we focus on pedestrian trajectory prediction. 

Early works on pedestrian trajectory prediction mainly focused on single-model prediction~\cite{socialforce_1995, sociallstm_2016_CVPR, socialattention_2018_ICRA}. However, these deterministic methods did not consider the inherent multi-modality of pedestrian movements. To address this problem, researchers have applied various generative methods to pedestrian trajectory prediction. For instance, Gupta~\etal~\cite{SocialGAN_2018_CVPR} utilized Generative Adversarial Network (GAN) for multi-model future generation, while Trajectron++~\cite{Trajectron++_2020_ECCV} used a Conditional Variational Autoencoder (CVAE) decoder. Nevertheless, GAN is hard to train and may face mode collapse, while VAE methods can produce unrealistic results. 

\begin{figure}[t]
    \scriptsize
    \setlength{\tabcolsep}{1.5pt}
    \centering
    \includegraphics[width=0.9\linewidth]{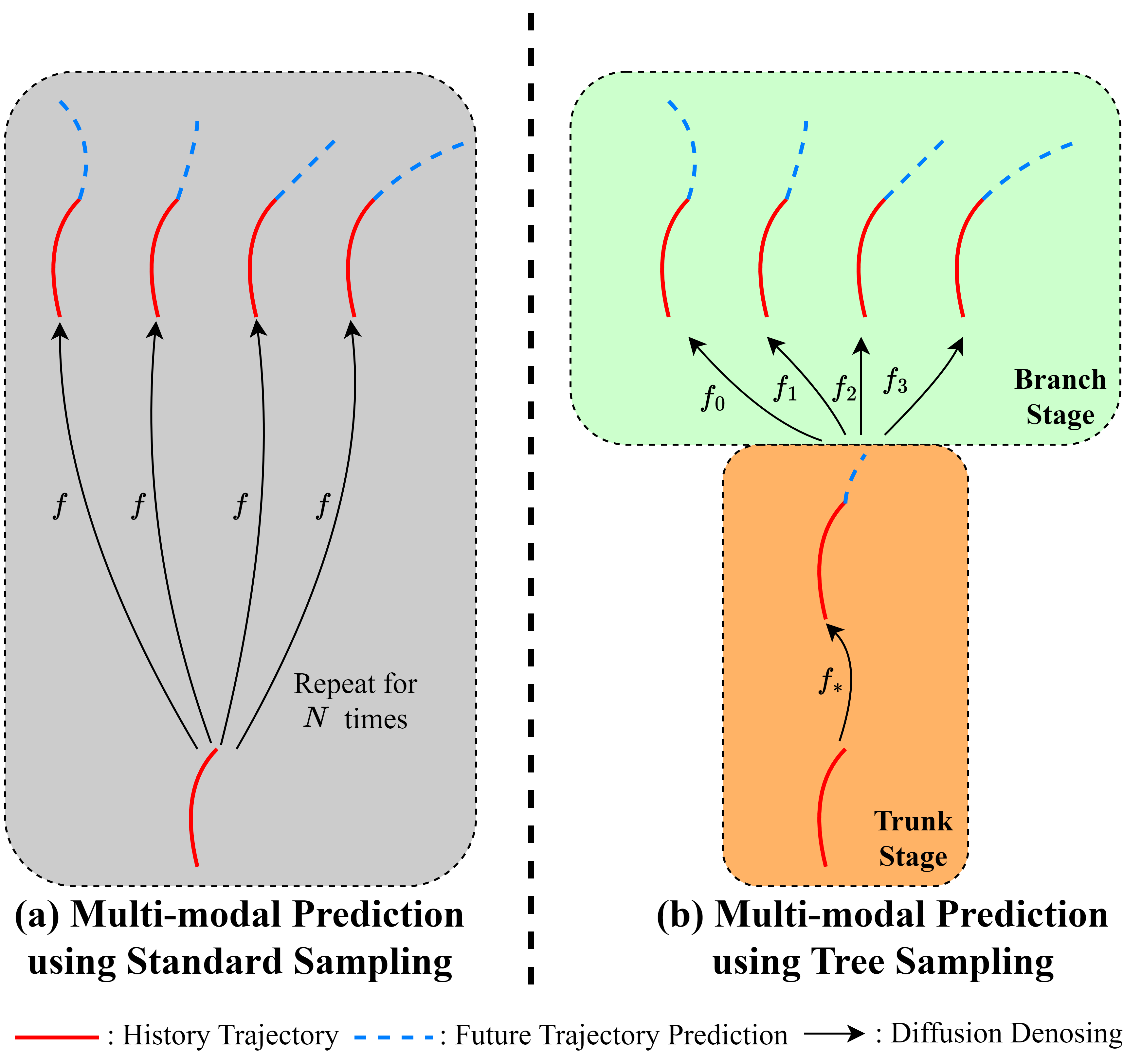}
    \\
    \caption{Comparison of standard sampling algorithm and our proposed tree sampling algorithm for multi-modal prediction. Since the trunk stage of our algorithm only needs to run once for multiple predictions, the total number of diffusion steps is fewer than the standard sampling algorithm, therefore accelerating the inference speed. }
    \label{fig:ts}
    \vspace{-14pt}
\end{figure}

Recently, denoising diffusion probabilistic models~\cite{DDPM_2020_Neurips} have achieved remarkable success in computer vision field~\cite{Classifier_2021_Neurips, labelefficient_2022_ICLR}. Many researchers have also adopted this generative method for robotics and autonomous driving applications. For example, Gu~\etal~\cite{MID_2022_CVPR} proposed MID based on diffusion models for pedestrian trajectory prediction. Following MID, LED~\cite{LED_2023_CVPR} also trained a denoising module using the same standard training schedule and incorporated a leapfrog initializer to skip many denoising steps to reduce inference time. Jiang~\etal~\cite{Motiondiffuser_CVPR_2023} proposed MotionDiffuser for multi-agent motion prediction and controllable trajectory synthesis.

Incorporating goal information into trajectory prediction models has been shown to improve the precision of predictions~\cite{DenseTNT_2021_ICCV, YNet_2021_ICCV, GoalSAR_2022_CVPRW, ADAPT_2023_ICCV}. Motivated by this, we combine goal estimation with diffusion models to leverage the "goal-driven" characteristic of human trajectory. Standard diffusion sampling algorithm used for multi-modal pedestrian trajectory prediction~\cite{MID_2022_CVPR} simply repeats the sampling procedure to generate multiple possible future trajectories, which could be rather time-consuming, as depicted in Fig. \ref{fig:ts}(a).

To address the aforementioned challenges, we propose a novel framework for multi-modal trajectory prediction in this work, as shown in Fig. \ref{fig:network}. In our proposed framework, the goal estimation module first estimates multiple possible goals to ensure the diversity of the predictions. These goals, along with the history motion information of the target agent, are then fed into the diffusion-based trajectory prediction module. To accelerate the diffusion sampling process, we design the two-stage tree sampling algorithm as Fig. \ref{fig:ts}(b) illustrates: In the trunk stage of the tree sampling, \textit{common feature} $\textbf{f}_*$ is first leveraged to generate a roughly denoised future trajectory. The trunk stage only needs to run once for arbitrary numbers of predictions, just like a tree having only one trunk. Then, the branch stage further denoises the trajectory from the trunk stage conditioned on \textit{diverse features} $\textbf{f}_n$ and generates corresponding trajectory predictions for each goal estimation. We leverage the powerful reconstruction ability of the deterministic diffusion model to further improve the accuracy of the prediction results. Experiment results show that our GBT framework achieves high prediction accuracy and fast inference speed. 
In summary, the main contributions of this paper are as follows:
\begin{itemize}
\item We propose a novel framework for multi-modal pedestrian trajectory prediction named \textbf{G}oal-guided \textbf{D}iffusion model with \textbf{T}ree \textbf{S}ampling (GDTS), which integrates goal estimation into conditional denoising diffusion model to improve the prediction performance.
\item We propose a two-stage tree sampling algorithm that increases the prediction accuracy and accelerates the inference speed without the requirement of additional networks. 
\item We conduct a series of experiments on various datasets, including ETH/UCY, Stanford Drone Dataset and intersection Drone Dataset, and the results attained demonstrate the state-of-the-art performance of our proposed framework in large-size datasets.
\end{itemize}

\section{RELATED WORKS}
\label{sec:related works}
\subsection{Goal-Guided Pedestrian Trajectory Prediction}
In general, pedestrian trajectory is goal-conditioned, as goals seldom change rapidly during the prediction horizon. Precise prediction of goals could reduce uncertainty and increase the accuracy of further trajectory prediction~\cite{YNet_2021_ICCV}. Several methods have been proposed to incorporate goal information into trajectory prediction. For exmaple, Dendorfer \etal~\cite{MG-GAN_2021_ICCV} trained multiple generators, and each generator is used to a different distribution associated with a specific mode. Mangalam \etal~\cite{PECNet_2020_ECCV} proposed a method that estimates goal points and generates multi-modal prediction using VAE conditioned on these estimated goal points. In \textsf{Y}-Net~\cite{YNet_2021_ICCV}, a U-Net architecture is employed for goal and trajectory prediction, where the predicted goal information is also fed into the trajectory prediction decoder. Chiara \etal~\cite{GoalSAR_2022_CVPRW} proposed a recurrent network, Goal-SAR, which predicts the position of each future time step in recurrent way based on the goal and all history information before this time step. NSP-SFM~\cite{NSPSFM_2022_ECCV} proposed by Yue~\etal also took advantage of goal estimation network with a similar structure. By predicting the parameter of goal attraction, inter-agent repulsion and environment repulsion in each timestep and generating future using CVAE, this NSP-SFM method reached SOTA performance. However, the recurrent prediction way could be time-comsuming. Moreover, it is unfair for NSP-SFM to tune hyperparameters separately for different scenes in ETH/UCY while other methods (including ours) use only one set of hyperparameters. In this paper, we also divide the pedestrian trajectory prediction task into goal estimation and trajectory prediction. 

\subsection{Diffusion Models and Application}
The diffusion probabilistic model was first proposed by Sohl-Dickstein~\etal~\cite{DPM_2015_ICML} and has developed rapidly since DDPM~\cite{DDPM_2020_Neurips} was proposed. It has achieved great success in various fields, such as computer vision~\cite{Classifier_2021_Neurips, labelefficient_2022_ICLR} and seq-to-seq models~\cite{DiffusionTimeseries_2021_ICML, DiffSeq_2022_Arxiv}. Diffusion models have also been adopted in robotics applications. For instance, Diffuser~\cite{Diffuser_2022_ICML} generates trajectories for robot planning and control using diffusion models. In Trace and Pace~\cite{Trace_2023_CVPR}, the diffusion model is used to produce realistic human trajectories. In the context of pedestrian trajectory prediction, MID~\cite{MID_2022_CVPR} was the first work to adopt the diffusion model for this task. Following the standard DDPM algorithm, a transformer acts as the diffusion network to generate multiple predictions conditioned on the history trajectory.

Despite its effectiveness in trajectory prediction, the diffusion model faces challenges due to the large number of denoising steps, which hampers real-time performance. 
To address this issue, various improvements have been proposed to accelerate the inference speed of the diffusion models~\cite{DiffusionSurvey_2022_Arxiv}. DDIM~\cite{DDIM_2021_ICLR} generalizes the forward process of the diffusion model as a non-Markovian process to speed up the sampling procedure. Salimans~\etal~\cite{PD_2022_ICLR} repeated knowledge distillation on a deterministic diffusion sampler to distill a new diffusion model with fewer sampling steps. Specific to multi-modal trajectory prediction, Mao~\etal~\cite{LED_2023_CVPR} proposes LED which trains an additional leapfrog initializer to accelerate sampling. Similar to our sampling algorithm, LED divides the diffusion sampling procedure into two stages and leverages an extra leapfrog initializer, which can estimate the roughly denoised distribution to skip the first stage. Instead of using an additional neural network, we use the same diffusion network to learn this distribution.
\begin{figure*}[th]
    \scriptsize
    \setlength{\tabcolsep}{1.5pt}
    \centering
    \includegraphics[width=1\linewidth]{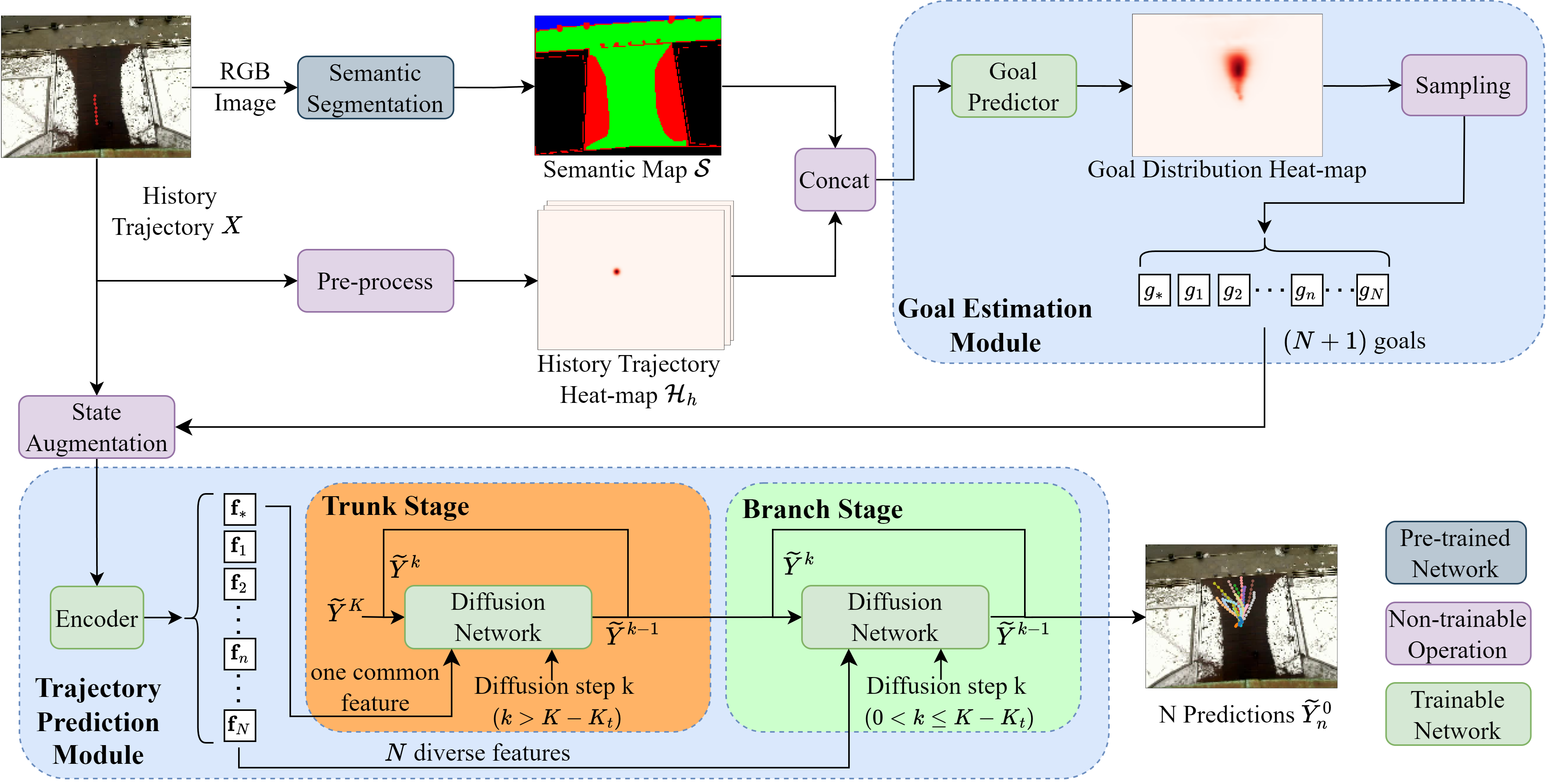}
    \\
    \caption{The architecture of our GDTS framework. GDTS consists of a goal estimation module and a diffusion-based trajectory prediction module. The goal estimation module predicts the goal distribution to estimate multiple goals, and the feature of history motion augmented with estimated goals information are then fed to the trajectory prediction module. The trajectory prediction module uses the two-stage tree sampling algorithm to denoise the trajectory from Gaussian noise $\widetilde{Y}^k$ to $\widetilde{Y}^{k-1}$ iteratively, conditioned on the inputted feature. During tree sampling, the common feature first acts as the guidance of denoising to generate a general initialization for further denoising in the trunk stage. Then in the branch stage, the diverse features "guide" the general initialization into different modalities to predict multi-modal future trajectories.}
    \label{fig:network}
    \vspace{-14pt}
\end{figure*}

\section{PROPOSED METHOD}
\label{sec:proposed method}
\subsection{Problem Formulation} 
The pedestrian trajectory prediction problem can be formulated as follows:
Given the current frame $t=0$, where the semantic map of the scene $\mathcal{S}$ and history positions of the targeted agent in the past $t_h$ frames $X = \{x_{t}\in \mathbb{R}^2| t = -t_h+1,-t_h+2,...,0 \}$ are available, the objective is to obtain $N$ possible future trajectory prediction of the agents in the next $t_f$ frames $\widetilde{Y}_n = \{\widetilde{y}_{t}\in \mathbb{R}^2| t = 1,2,...,t_f \}$. The notation with tilde represents the predicted result. 

Our proposed framework divides the overall trajectory prediction task into goal estimation and trajectory prediction, as illustrated in Fig. \ref{fig:network}. First, a network is applied to predict the probability distribution heat-map of the goal position. Then, a set of possible goals of the target agent is sampled from this distribution heat-map. Extracting the feature of these goal estimations and history information as guidance, a diffusion denoising network using the two-stage tree sampling algorithm is applied to get the multi-modal future trajectory prediction $\widetilde{Y}_n$ of the target agent. In the following part of this section, we will introduce the proposed framework in detail. Furthermore, we will introduce the loss function and the training scheme. 

\subsection{Goal Estimation Module}
\label{subsec:goal pred}
Considering the state-of-the-art performance of \textsf{Y}-Net~\cite{YNet_2021_ICCV} and Goal-SAR\cite{GoalSAR_2022_CVPRW} on pedestrian trajectory prediction, we adapt their goal predictor for goal estimation. 
The goal predictor takes the semantic map and history trajectory as input. Since our work focuses on trajectory prediction, we assume that the semantic network is pre-trained to obtain semantic map $\mathcal{S}$~$\in$~$\mathbb{R}^{H\times W\times C}$, where $H$ and $W$ are the height and width of the input image, and $C$ is the number of semantic classes.
Since we use a probability heat-map to describe the goal distribution, to keep the consistency of the input and output, pre-processing is needed before goal estimation to convert the inputted history positions of the frame $x_t$ to a 2D Gaussian probability distribution heat-map $\mathcal{H}_t$~$\in$~$\mathbb{R}^{H\times W}$ for each frame, and the highest probability is assigned to the ground truth position $x_t$. These position heat-maps are then stacked together to form a trajectory heat-map $\mathcal{H}_{h}$~$\in$~$\mathbb{R}^{H\times W\times t_h}$.
Finally, $\mathcal{H}_{h}$ and $\mathcal{S}$ are concatenated and fed into a U-Net architecture network to predict the future trajectory heat-map $\widetilde{\mathcal{H}}_{f}\in$~$\mathbb{R}^{H\times W\times t_f}$, and the last channel represents the position distribution of the last frame $t={t_f}$, i.e., the goal distribution. 
By sampling from the goal probability distribution heat-map, $N$ estimated goals $g_n, n=1,..., N$ are obtained, referred to as \textit{diverse goals}. Meanwhile, the position with the highest probability is selected as the \textit{common goal} $g_*$. Therefore, a total of $N+1$ estimated goals $G$ are generated. 

\begin{algorithm}[hbtp]
\label{algo1}
\KwData{\textit{Estimated goals} $G$, history trajectory $X$, tree sampling parameters $K$, $K_I$, $K_{t}$, $N$, $\eta$}
\For{$g \in G$ \tcp{Input augmentation}}{$V = \mathrm{d}_t X, A = \mathrm{d}_t V$ \\
$D = \{x_t - g |t=-t_h+1, -t_h+2, ..., 0\}$ \\
$\hat{X} = concat(D, X, V, A)$\\$f=LSTM(\hat{X})$}
$x_K\sim\mathcal{N}(0,\mathbf{I})$ \\
\For{$k = K, ..., K-K_{t}+1$ \tcp{Trunk stage}}{ $\widetilde{Y}^{k-1}=\frac{1}{\sqrt{\alpha_k}}(\widetilde{Y}^k - \frac{1-\alpha_k}{\sqrt{1-\bar{\alpha}_k}}\epsilon(\widetilde{Y}^k,k,\mathbf{f}_*))$}
$\sigma = \sqrt{\eta \frac{1-\bar{\alpha}_{k-1}}{1-\bar{\alpha}_k} (1-\frac{\bar{\alpha}_k}{\bar{\alpha}_{k-1}})}$ \\
$K_{b} = (1 - \frac{K_{t}}{K})K_I $ \\ 
\For{$n=1,...N$ \tcp{Branch stage}}{$\widetilde{Y}_n^{K_b} = \widetilde{Y}^{K-K_t}$\\ \For{$k = K_{b}, ..., 1$}{$\mathbf{z}\sim\mathcal{N}(0,\mathbf{I})$ if $k>1$, else $\mathbf{z}=0$\\$\widetilde{Y}_n^{k-1}= \sqrt{\frac{\bar{\alpha}_{k-1}}{\bar{\alpha}_{k}}} \widetilde{Y}_n^k + (\sqrt{1 - \bar{\alpha}_{k-1} - \sigma^2} - \sqrt{\bar{\alpha}_{k-1} \frac{1-\bar{\alpha}_k}{\bar{\alpha}_k}}) \epsilon(\widetilde{Y}_n^k,k,\mathbf{f}_n) + \sigma \mathbf{z}$}}
\Return $N$ predictions $\widetilde{Y}_n, n=1,...,N$
\caption{Diffusion-Based Trajectory Prediction with Tree Sampling Using Estimated Goals}
\end{algorithm}
\subsection{Trajectory Prediction Module}
\label{subsec:traj generation}
The trajectory prediction module receives the estimated goals $G$ and the history trajectory $X$ as input. Following previous works \cite{Trajectron++_2020_ECCV, MID_2022_CVPR}, we augment the history state $X$ with velocity and acceleration. Additionally, we integrate the goal information into the state by concatenating the vector from the position of this frame $x_t$ to the goal $g \in G$, as shown in lines 3-6 of Algorithm \ref{algo1}.
After obtaining the augmented state $\hat{X}$ for each goal estimation, an LSTM encoder is trained to extract the feature $\textbf{f}$ of the state. Similarly, We refer to the feature $\textbf{f}_n$ obtained using $g_n$ as \textit{diverse feature}, and the feature $\textbf{f}_*$ obtained using $g_*$ as \textit{common feature}. And $\textbf{f}$ then serves as the guidance of the diffusion.
During the forward process of diffusion, noise is added to the ground truth trajectory $Y^0$, and this noise addition is repeated for $K$ times to obtain a series of noisy future trajectories $\{Y^1, ..., Y^K\}$; While in step $k$ of the reverse denoising process, the diffusion network predicts a noise conditioned on $k$ and $\textbf{f}$. This predicted noise is then used to denoise the noisy future trajectory $\widetilde{Y}^k$ to $\widetilde{Y}^{k-1}$. With the iterative denoising process, the future trajectory prediction $\widetilde{Y} = \widetilde{Y}^0$ can be gradually reconstructed from a Gaussian distribution $Y^K\sim\mathcal{N}(0,\mathbf{I})$. Standard sampling algorithms (DDPM~\cite{DDPM_2020_Neurips} and DDIM~\cite{DDIM_2021_ICLR}) repeat the whole sampling procedure to generate multiple modalities, which could be time-consuming. To alleviate this drawback, we propose tree sampling, as shown in Algorithm \ref{algo1}. Our tree sampling algorithm is designed especially for multi-modal prediction and is divided into two stages: the trunk stage and the branch stage. The total number of DDPM diffusion steps is $K$, and the number of DDIM diffusion steps is $K_I$. The trunk stage consists of $K_t$ steps, and the branch stage consists of $K_b$ steps. In the trunk stage (lines 10-12), we begin by utilizing the \textit{common feature} $\textbf{f}_*$ to denoise the trajectory. By applying a variant of DDPM which removes the noise term $\sigma_t\textbf{z}$ (we call this variant deterministic-DDPM or d-DDPM in short), the deterministic denoised result of the trunk stage $\widetilde{Y}^{K-K_t}$ is obtained, which could serve as general initialization for further denoising conditioned on different \textit{diverse features} $\textbf{f}_n$, i.e., the trunk which links to different branches of this tree.  
In the subsequent branch stage (lines 15-22), different $\textbf{f}_n$ are used for refinement of $\widetilde{Y}^{K-K_t}$ to obtain final multi-modal predictions $\widetilde{Y}^{0}_n$. Since the branch stage needs to be run multiple times for different modalities, we apply DDIM in this stage to increase the inference speed. Experiments demonstrate our combination scheme generates more accurate results in real-world datasets.

\subsection{Training and Loss Function}
Our training scheme consists of two stages to train two modules separately. First, the goal estimation module is trained solely using Binary Cross-Entropy loss, which measures the dissimilarity between the ground truth of future trajectory heat-map $\mathcal{H}_f$ and predicted heat-map $\widetilde{\mathcal{H}}_f$:
\begin{equation}\label{eqn:1}
\begin{aligned}
\mathcal{L}_{goal} = BCE(\mathcal{H}_f, \widetilde{\mathcal{H}}_f)
\end{aligned}
\end{equation}
\begin{equation}\label{eqn:2}
\begin{aligned}
\mathcal{L}_{1} = \mathcal{L}_{goal}
\end{aligned}
\end{equation}
In the second stage, we train the goal estimation module together with the trajectory prediction module. The training objective of the trajectory prediction module is to learn a distribution $\epsilon\sim\mathcal{N}(0,\mathbf{I})$ for each step $k$ of diffusion, following the regular setting of diffusion models \cite{DDPM_2020_Neurips, DDIM_2021_ICLR}:
\begin{equation}\label{eqn:3}
\begin{aligned}
\mathcal{L}_{traj} = \mathbb{E}||\epsilon_{\theta}(k, Y^k, \textbf{f}) - \epsilon||
\end{aligned}
\end{equation}
\begin{equation}\label{eqn:4}
\begin{aligned}
\mathcal{L}_{2} = \lambda \mathcal{L}_{goal} + \mathcal{L}_{traj}
\end{aligned}
\end{equation}

\section{RESULTS}

\subsection{Experiments and Datasets}
\label{sec:experiments}
To evaluate the performance of our model, we conducted experiments on three public real-world pedestrian datasets: the ETH/UCY dataset, the Stanford Drone Dataset (SDD) and the intersection Drone Dataset (inD). These datasets capture the movement of pedestrians from a bird's-eye view perspective and positions are manually annotated. The sample rate of datasets is 2.5 FPS. Given the pedestrian position of the last $t_h = 8$ frames, the task is to predict the trajectory of the following $t_f = 12$ frames. 
\begin{table*}[t]
\caption{\textbf{Quantitative results on ETH/UCY dataset.} \dag indicates the model after the data leakage issue is fixed.} 
\centering
\resizebox{2\columnwidth}{!}{
\begin{tabular}{l|c c |c c |c c |c c |c c |c c}
    \toprule
    ~&\multicolumn{2}{c|}{\textbf{ETH}} & \multicolumn{2}{c|}{\textbf{HOTEL}} &  \multicolumn{2}{c|}{\textbf{UNIV}} &  \multicolumn{2}{c|}{\textbf{ZARA1}} & \multicolumn{2}{c|}{\textbf{ZARA2}} & \multicolumn{2}{c}{\textbf{AVG}} \\
    \cmidrule{2-13}
    &\multicolumn{1}{c}{\textbf{ADE$_{20}$}} & \multicolumn{1}{c|}{\textbf{FDE$_{20}$}} &
    \multicolumn{1}{c}{\textbf{ADE$_{20}$}} &  \multicolumn{1}{c|}{\textbf{FDE$_{20}$}} &
    \multicolumn{1}{c}{\textbf{ADE$_{20}$}} &  \multicolumn{1}{c|}{\textbf{FDE$_{20}$}} &
    \multicolumn{1}{c}{\textbf{ADE$_{20}$}} &  \multicolumn{1}{c|}{\textbf{FDE$_{20}$}} &
    \multicolumn{1}{c}{\textbf{ADE$_{20}$}} &  \multicolumn{1}{c|}{\textbf{FDE$_{20}$}} &
    \multicolumn{1}{c}{\textbf{ADE$_{20}$}} &  \multicolumn{1}{c}{\textbf{FDE$_{20}$}}
    
    \\
    \cmidrule{1-13}
    Goal-GAN~\cite{Goal-GAN_2020_ACCV}  & 0.59&1.18 & 0.19&0.35 &  0.60&1.19 & 0.43&0.87 & 0.32&0.65 & 0.43&0.85\\
    MG-GAN~\cite{MG-GAN_2021_ICCV}  & 0.47&0.91 & 0.14&0.24 & 0.54&1.07 & 0.36&0.73 & 0.29&0.60 &  0.36&0.71\\
    PECNet~\cite{PECNet_2020_ECCV}  & 0.54&0.87 & 0.18&0.24 & 0.35&0.60 & 0.22&0.39 & 0.17&0.30 & 0.29&0.48\\
    \textsf{Y}-net~\cite{YNet_2021_ICCV}  & 0.28&0.33 & 0.10&0.14 & 0.24&0.41 & 0.17&0.27 & 0.13&0.22 & 0.18&0.27 \\
    Goal-SAR~\cite{GoalSAR_2022_CVPRW} & 0.28&0.39 & 0.12&0.17 & 0.25&0.43 & 0.17&0.26 &  0.15&0.22 &  0.19&0.29\\
    \cmidrule{1-13}
    MID\dag~\cite{MID_2022_CVPR}  & 0.57&0.93 & 0.21&0.33 & 0.29&0.55 & 0.28&0.50 & 0.20&0.37 &  0.31&0.54\\
    LED~\cite{LED_2023_CVPR} & 0.39&0.58 & 0.11&0.17 & 0.26&0.43 & 0.18&0.26 & 0.13&0.22 &  0.21&0.33\\
    SingularTrajectory~\cite{Singulartrajectory_CVPR_2024} & 0.35&0.42 &0.13&0.19 &0.25&0.44 &0.19&0.32 &0.15&0.25 &0.21&0.32 \\
    \cmidrule{1-13}
    GDTS (Ours) & 0.31&0.48 & 0.13&0.18 & 0.27&0.49 & 0.19&0.29 & 0.15&0.24 &  0.21&0.33\\
    \bottomrule
\end{tabular}
}
\label{table:eth_ucy}
\end{table*}

\begin{table}[t]
\caption{\textbf{Quantitative results on Stanford Drone Dataset (SDD).} Bold number indicates the \textbf{best}.} 
\centering
\resizebox{0.6\columnwidth}{!}{
\begin{tabular}{c|c|c}
    \toprule
    \textbf{Method} & \textbf{ADE$_{20}$} & \textbf{FDE$_{20}$}\\
    \cmidrule{1-3}
    Goal-GAN~\cite{Goal-GAN_2020_ACCV} & 12.20 & 22.10\\
    MG-GAN~\cite{MG-GAN_2021_ICCV} & 13.60 & 25.80\\
    PECNet~\cite{PECNet_2020_ECCV} & 9.96 & 15.88\\
    \textsf{Y}-net~\cite{YNet_2021_ICCV} & 7.85 & 11.85\\
    Goal-SAR~\cite{GoalSAR_2022_CVPRW} & 7.75 & 11.83\\
    \cmidrule{1-3}
    MID~\cite{MID_2022_CVPR} & 7.61 & 14.30\\
    LED~\cite{LED_2023_CVPR} & 8.48 & 11.66\\
    \cmidrule{1-3}
    GDTS (Ours) & \textbf{7.42} & \textbf{11.57}\\
    \bottomrule
\end{tabular}
}
\label{table:SDD}
\end{table}

\textbf{Datasets}
The ETH/UCY~\cite{ETH_2009_ICCV, UCY_2007} dataset consists of 5 scenes: ETH, Hotel, Univ, Zara1, and Zara2. Following the common leave-one-scene-out strategy in previous work~\cite{SocialGAN_2018_CVPR, YNet_2021_ICCV}, we use four scenes for training and the remaining one for testing. SDD~\cite{SDD_2016_ECCV} is a large-scale dataset that contains 11,216 unique pedestrians on the university campus. We use the same dataset split as several recent works \cite{GoalSAR_2022_CVPRW, PECNet_2020_ECCV}: 30 scenes are used for training, and the remaining 17 scenes are used for testing. inD~\cite{ind_IV_2020} includes 4 different intersection scenes in German and we follow the same evaluation protocol applied in ~\cite{GoalSAR_2022_CVPRW, ACVRNN_2021, gsgformer_2023gsgformer}, where only pedestrian trajectories are retained and split into train, validation, and test sets by a 70\%/10\%/20\% portion.

\textbf{Evaluation Metrics}
Average Displacement Error (ADE) measures the average Euclidean distance between the ground truth and the predicted positions of the entire future trajectory, while Final Displacement Error (FDE) considers only the Euclidean distance between the final positions. Since our model generates multi-model predictions for stochastic future, we report the best-of-$N$ ADE and FDE of the predictions, denoted as ADE$_N$ and FDE$_N$.

\textbf{Implementation Details}
All the experiments were conducted on an NVIDIA 3080Ti GPU with PyTorch implementation. 
The goal predictor is implemented using a 5-layer U-Net as the backbone. The architecture of the diffusion network is a three-layer transformer encoder similar to MID~\cite{MID_2022_CVPR}, and readers can refer to the original paper for more details.
We adopt a batch size of $64$. An Adam optimizer is employed with an initial learning rate of $10^{-3}$, and exponential annealing is applied for the learning rate. The goal estimation module is trained for $150$ epochs first and then trained with trajectory prediction module for $250$ epochs. The DDPM diffusion step $K$ is $100$, the DDIM diffusion step $K_I$ is $20$, and the trunk step $K_{t}$ is $30$. 
Additionally, we utilize Test-Time-Sampling-Trick (TTST) in our model to improve the accuracy of goal estimation \cite{YNet_2021_ICCV}, and the number of TTST samples $N_{ttst}$ is selected to be $1000$. The goal estimation loss coefficient $\lambda=20$ and The final number of predictions $N=20$.
\begin{table}[!tbp]
\caption{\textbf{Quantitative results on Intersection Drone Dataset (inD).} Bold number indicates the \textbf{best}. * indicates the result reported in \cite{GoalSAR_2022_CVPRW}} 
\centering
\resizebox{0.6\columnwidth}{!}{
\begin{tabular}{c|c|c}
    \toprule
    \textbf{Method} & \textbf{ADE$_{20}$} & \textbf{FDE$_{20}$}\\
    \cmidrule{1-3}
    Social-GAN~\cite{SocialGAN_2018_CVPR} & 0.48 & 0.99\\
    ST-GAT~\cite{STGAT_2019_ICCV} & 0.48 & 1.00\\
    AC-VRNN~\cite{ACVRNN_2021} & 0.42 & 0.80\\
    GSGFormer~\cite{gsgformer_2023gsgformer} & \textbf{0.30} & 0.55 \\
    \textsf{Y}-net*~\cite{YNet_2021_ICCV} & 0.34 & 0.56\\
    Goal-SAR~\cite{GoalSAR_2022_CVPRW} & 0.31 & 0.54\\
    \cmidrule{1-3}
    GDTS (Ours) & \textbf{0.30} & \textbf{0.53}\\
    \bottomrule
\end{tabular}
}
\label{table:inD}
\end{table}

\begin{figure*}[t]
    \scriptsize
    \setlength{\tabcolsep}{1.5pt}
    \centering
    \includegraphics[width=1\linewidth]{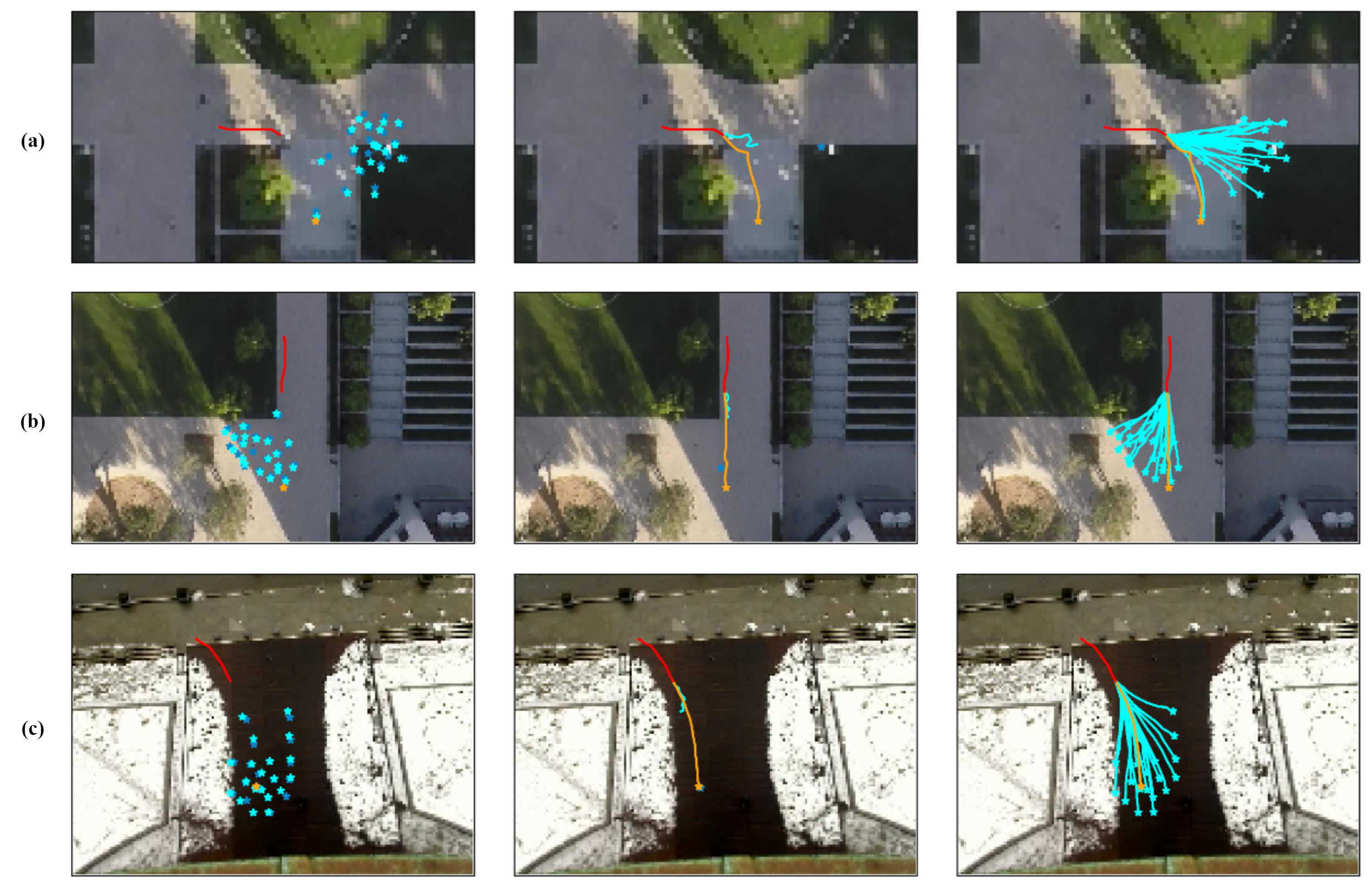}
    \\
    \caption{Visualization of history trajectory and future prediction of three agents in different scenes on the ETH/UCY and SDD. The observed trajectory is in red, the goal estimations obtained from the goal estimation module are in blue, the final predictions are in cyan, and the ground truth goal and trajectory are in yellow. The goals are highlighted as star points.}
    \label{fig:qual_results}
    \vspace{-14pt}
\end{figure*}

\subsection{Quantitative Results}
\textbf{Baseline}
We compare GDTS to different methods: Goal-based methods include Goal-GAN~\cite{Goal-GAN_2020_ACCV}, MG-GAN~\cite{MG-GAN_2021_ICCV}, PECNet~\cite{PECNet_2020_ECCV}, \textsf{Y}-net~\cite{YNet_2021_ICCV}, Goal-SAR~\cite{GoalSAR_2022_CVPRW}. We do not include NSP-SFM~\cite{NSPSFM_2022_ECCV} here due to its unfair hyperparameters tuning for each scene in ETH/UCY. MID~\cite{MID_2022_CVPR}, LED~\cite{LED_2023_CVPR} and SingularTrajectory~\cite{Singulartrajectory_CVPR_2024} are methods which leverage diffusion model for pedestrian trajectory prediction. Specifically for inD, besides goal-based methods, we also compare our model with some SOTA methods: Social-GAN~\cite{SocialGAN_2018_CVPR}, ST-GAT~\cite{STGAT_2019_ICCV}, AC-VRNN~\cite{ACVRNN_2021} and GSGFormer~\cite{gsgformer_2023gsgformer}, which are all training with the same dataset split as our model. Note that recent works with different evaluation protocal~\cite{hierarchical_TVT_2024} are not included for fair comparison.

\begin{table}[t]
\caption{\textbf{Ablation study on sampling algorithm of diffusion and $K_{t}$}. The last row is the setting of our final reported results. Bold number indicates the \textbf{best}.}
\centering
\resizebox{!}{1.65cm}{
\begin{tabular}{c | c | c | c | c}
    \toprule
    \makecell[c]{Sampling\\Algorithm}   & \textbf{$K_t$}  &\textbf{ADE$_{20}$} & \textbf{FDE$_{20}$} & \makecell[c]{Inference\\(ms)}\\
    \cmidrule{1-5}
    MID(DDPM) & - & 7.61 & 14.30 & $\sim$139 \\
    \cmidrule{1-5}
     DDPM & - & 7.76 & 11.91 & $\sim$65\\
     DDIM & - & 7.91 & 12.01 & $\sim$24\\ 
     TS & 5 & 7.61 & 11.88 & $\sim$24\\
     TS & 50 & 7.56 & 12.08 & $\sim$\textbf{21}\\
     \cmidrule{1-5}
    TS (Ours) & 30 & \textbf{7.42} & \textbf{11.57} & $\sim$24\\
    \bottomrule
\end{tabular}
}
\label{table:ablation_1}
\end{table}

\textbf{Discussion} 
Results in Table \ref{table:eth_ucy} show that our model achieves comparable accuracy in the ETH/UCY dataset. Furthermore, results of experiments on the SDD are reported in Table \ref{table:SDD}. With the combination of explicit guide guidance and diffusion denoising, our model achieves the best ADE$_{20}$ and FDE$_{20}$ among all compared methods. Results in inD shown in Table \ref{table:inD} are in line with the previous table, further confirming the effectiveness of GDTS. Since SDD and inD are substantially larger than the ETH/UCY dataset, we argue that our method is more suitable for datasets with larger sizes.

\begin{table}[t]
\caption{\textbf{Ablation study on combination scheme of tree sampling algorithm}. The last row is the setting of our final reported results. Bold number indicates the \textbf{best}.} 
\centering
\resizebox{\columnwidth}{!}{
\begin{tabular}{c|c|c|c|c|c|c}
    \toprule
    \makecell[c]{\textbf{Trunk}\\ \textbf{Stage}} &\makecell[c]{\textbf{Branch}\\ \textbf{Stage}}  & \textbf{$K_t$}  & \textbf{$K_b$} &\textbf{ADE$_{20}$} & \textbf{FDE$_{20}$} & \makecell[c]{Inference\\(ms)}\\
    \cmidrule{1-7} 
     d-DDPM & DDPM & 30 & 70 & 7.74 & 11.83 & $\sim$65\\
     d-DDPM& d-DDPM & 30 & 70 & \textbf{7.36} & 11.63 & $\sim$65\\
     DDIM & DDIM & 6 & 14 & 8.02 & 12.62 & $\sim$\textbf{21}\\
     \cmidrule{1-7}
     d-DDPM & DDIM & 30 & 14 & 7.42 & \textbf{11.57} & $\sim$24\\
    \bottomrule
\end{tabular}
}
\label{tab:ablation_2}
\end{table}

\subsection{Ablation Study}
\textbf{Sampling Algorithm of Diffusion}
To investigate the influence of different sampling algorithms, we replace tree sampling (TS) with DDPM and DDIM, and conduct the experiment on SDD. We report the average inference time for one agent in addition to the ADE$_{20}$ and FDE$_{20}$. Since LED~\cite{LED_2023_CVPR} and SingularTrajectory~\cite{Singulartrajectory_CVPR_2024} do not have an official implementation nor reported inference time on SDD, we only include MID which uses DDPM for comparison. The results in Table \ref{table:ablation_1} verify that our proposed tree sampling algorithm performs the best and has a shorter inference time than DDPM and DDIM, indicating the superiority of our sampling algorithm. Compared with MID, our GDTS method reduces the inference time from 139 ms to around 25 ms. 

\textbf{Trunk Step of Tree Sampling}
We also explore the effect of the trunk step $K_{t}$. As shown in Table \ref{table:ablation_1}, $K_{t}$ does not significantly influence the inference time since we use DDIM in the branch stage which already has a fast sampling speed. However, the accuracy of our method decreases when $K_{t}$ is either too large or too small. When $K_{t}$ is large, small $K_{b}$ makes it challenging for the branch stage to drive the roughly denoised trajectory to various modalities, limiting the diversity of the results. When $K_{t}$ is small, the branch stage that uses DDIM undertakes most of the denoising task, without obtaining enough guidance from the trunk stage, which degrades the performance. 

\textbf{Combination scheme of Tree Sampling}
We compare different combinations of DDPM, d-DDPM, and DDIM in tree sampling algorithm to verify the effectiveness of our scheme. Note that the result of the trunk stage must be deterministic, hence only d-DDPM and DDIM ($\eta=0$) can be used in the trunk stage. Results in Table \ref{tab:ablation_2} demonstrate that using d-DDPM in both stages gets the lowest ADE$_{20}$ and FDE$_{20}$, probably because d-DDPM focuses on reconstruction rather than generation, i.e., accuracy rather than diversity. However, the inference time is much longer than our scheme, and our scheme achieves a good balance between prediction accuracy and inference speed.

\subsection{Qualitative Results}
Fig. \ref{fig:qual_results} presents the qualitative results of our GDTS method on the ETH/UCY and SDD datasets. The target pedestrian in the scene (a) turns right while moving straight in the scene (b) and (c). From the left column, it can be observed that the goals from the final prediction (in cyan) are closer to the ground truth (in yellow) compared to the goal estimations obtained from the goal estimation module (in blue), indicating that the diffusion models can improve FDE. The middle column shows the \textit{common goal} $g_*$ and the roughly denoised trajectory generated by the trunk stage, demonstrating that the \textit{common goal} can guide the denoised trajectory growing toward the correct direction. The right column shows the final prediction results, demonstrating the ability of GDTS to generate future trajectory predictions with different modalities.

\section{CONCLUSIONS}
In conclusion, GDTS is introduced in this work for multi-modal pedestrian trajectory prediction. The proposed development is a framework that integrates goal estimation and the diffusion probabilistic model. Our novel tree sampling algorithm leverages \textit{common feature} for multiple modalities generation to accelerate inference speed. Experiments on real-world datasets demonstrate that our GDTS method can predict multiple scene-compliant trajectories and achieve state-of-the-art performance with real-time inference speed in real-world datasets. 
Nevertheless, the interaction between agents is currently not considered in this work, yet it is crucial for human motion prediction. Predicting additional probabilistic score for each modality could help with downstream decision-making and planning. In this sense, we will explore the integration of multi-agent interactions into our framework.








\renewcommand*{\bibfont}{\footnotesize}
\printbibliography

\end{document}